\documentclass[conference]{IEEEtran}
\IEEEoverridecommandlockouts

\usepackage{cite}
\usepackage{amsmath,amssymb,amsfonts}
\usepackage{algorithmic}
\usepackage{graphicx}
\usepackage{textcomp}
\usepackage{xcolor}
\usepackage{bibentry}
\usepackage{booktabs} 
\usepackage{multirow} 
\definecolor{deepgreen}{rgb}{0.0, 0.5, 0.0}

\def\BibTeX{{\rm B\kern-.05em{\sc i\kern-.025em b}\kern-.08em
    T\kern-.1667em\lower.7ex\hbox{E}\kern-.125emX}}
\begin{document}
\title{Multi-modal Knowledge Graph Generation with Semantics-enriched Prompts\\
}

\author{
  \IEEEauthorblockN{\normalsize Yajing Xu\textsuperscript{1}, Zhiqiang Liu\textsuperscript{1}, Jiaoyan Chen\textsuperscript{2}, Mingchen Tu\textsuperscript{1}, Zhuo Chen\textsuperscript{1}, Jeff Z. Pan\textsuperscript{3}\\ Yichi Zhang\textsuperscript{1}, Yushan Zhu\textsuperscript{1}, Wen Zhang\textsuperscript{1}$^\dagger$, Huajun Chen\textsuperscript{1}$^\dagger$}
  \IEEEauthorblockA{
    \textit{\textsuperscript{1} Zhejiang University, Hangzhou, China} \\
    \textit{\textsuperscript{2} Department of Computer Science, University of Manchester, United Kingdom} \\
    \textit{\textsuperscript{3} School of Informatics, The University of Edinburgh, United Kingdom} \\
    \small \{yajingxu, zhiqiangliu\}@zju.edu.cn, jiaoyan4ai@gmail.com, \{mingchentz, zhuo.chen\}@zju.edu.cn \\
    \small j.z.pan@ed.ac.uk, \{zhangyichi2022, yushanzhu, zhang.wen, huajunsir\}@zju.edu.cn
  }
}

\maketitle

\def\thefootnote{$\dagger$}\footnotetext{Corresponding Author.}

\begin{abstract}
Multi-modal Knowledge Graphs (MMKGs) have been widely applied across various domains for knowledge representation. However, the existing MMKGs are significantly fewer than required, and their construction faces numerous challenges, particularly in ensuring the selection of high-quality, contextually relevant images for knowledge graph enrichment. To address these challenges, we present a framework for constructing MMKGs from conventional KGs. Furthermore, to generate higher-quality images that are more relevant to the context in the given knowledge graph, we designed a neighbor selection method called \textbf{V}isualizable \textbf{S}tructural \textbf{N}eighbor \textbf{S}election (VSNS). This method consists of two modules: \textbf{V}isualizable \textbf{N}eighbor \textbf{S}election (VNS) and \textbf{S}tructural \textbf{N}eighbor \textbf{S}election (SNS). The VNS module filters relations that are difficult to visualize, while the SNS module selects neighbors that most effectively capture the structural characteristics of the entity. To evaluate the quality of the generated images, we performed qualitative and quantitative evaluations on two datasets, MKG-Y and DB15K. The experimental results indicate that using the VSNS method to select neighbors results in higher-quality images that are more relevant to the knowledge graph.
\end{abstract}

\section{Introduction}
Multi-modal Knowledge Graphs (MMKGs) are advanced Knowledge Graphs (KGs) that integrate both semantic and visual information, providing a comprehensive description of entities and making them applicable to a broad range of environments \cite{chen2024knowledge,DBLP:journals/tgdk/PanRKSCDJO0LBMB23}. 
However, the construction of MMKGs faces numerous challenges. Common approaches include enriching existing KGs with multimedia data from repositories such as Wikipedia or retrieving images through search engines. Nevertheless, both repository-based and search engine-based methods struggle to ensure the selection of high-quality, contextually relevant images for KG enrichment.

In recent years, the rapid advancement of generative AI has made it possible to generate high-quality images that are nearly indistinguishable from the real world \cite{DBLP:conf/nips/HedlinSMIKTY23}. This capability offers a novel approach for MMKG construction: Generating images for entities in conventional KGs.
However, existing methods that leverage these generative models \cite{DBLP:conf/nips/HoJA20} often rely on manually crafted prompts, which is time-consuming and becomes increasingly impractical as the size of the enriched database grows. To address this limitation, an automated method is essential for generating entity-specific prompts. The most straightforward approach is to use the entity's name as the prompt. However, this often fails to generate images that accurately capture the entity's characteristics, particularly when the generative model's pre-trained data lacks sufficient information about the entity. To overcome this issue, an intuitive solution is to incorporate information from the entity's neighbors. Nevertheless, directly inputting all neighbor information into the generative model is impractical due to potential limitations in the model's input size. Moreover, many neighbors may represent redundant or similar connections, making it unnecessary to include all of them for image generation. Instead, selecting a subset of the most relevant neighbors is more efficient.

Recent work \cite{DBLP:conf/wikidata/AhmadCEMRZM23} highlights the importance of triple content in generating high-quality images. Specifically, the study notes that while the number of triples or unique relations does not significantly impact image quality, the object within triples play a crucial role. For instance, the properties like "instance of" for fictional characters has a substantial impact on the quality of generated images. This finding underscores the significance of selecting relations that are not only relevant but also visually descriptive for image generation.

In this paper, we propose a novel neighbor selection method named \textbf{V}isualizable \textbf{S}tructural \textbf{N}eighbor \textbf{S}election (VSNS), which consists of two modules: \textbf{V}isualizable \textbf{N}eighbor \textbf{S}election (VNS) and \textbf{S}tructural \textbf{N}eighbor \textbf{S}election (SNS). The VNS module selects relations that are easier to visualize, ensuring that the generated images are both contextually relevant and visually meaningful. The SNS module, on the other hand, focuses on selecting neighbors that best capture the structural characteristics of the entity. By combining these modules, VSNS ensures that the selected neighbors provide the most informative and visually descriptive content for image generation. Using the selected neighbors, we employ predefined instructions to guide a language model in generating semantics-enriched prompts. These prompts are then used by a generative model to produce images for the entity. 

To evaluate the quality of the generated images, we conducted a manual assessment, ranking both generated and real images across multiple dimensions. Additionally, we used two widely used metrics, FID (Fréchet Inception Distance) and CLIPscore, to quantitatively assess the images. Finally, to explore the impact of the generated images on the knowledge graph, we tested their performance in inference tasks on the MMKG.

Our contributions can be summarized as follows:
\begin{itemize}
    \item We propose a comprehensive framework for constructing MMKGs from conventional KGs. The framework consists of three modules: Visualizable Structural Neighbor Selection, Semantics-enriched Prompts, and Image Generation. This framework effectively translates KGs into MMKGs.
    \item We introduce a novel neighbor selection method, VSNS, composed of VNS and SNS. VSNS enables the generation of images that are more relevant to the entities and better aligned with the knowledge graph.   
    \item We conduct a thorough evaluation of the generated images across two datasets, empirically validating the effectiveness of our methodology. 
\end{itemize}

\section{Preliminary}
A knowledge graph (KG) is defined as $\mathcal{G}=\{\mathcal{E}, \mathcal{R}, \mathcal{T}\}$. $\mathcal{E}$ is the set of entities that includes individuals like people, places, and organizations. $\mathcal{R}$ is the set of relations, which consists of relations between entities, such as \textit{hasFriend} and \textit{locatedIn}. $\mathcal{T} = \{ (h, r, t) | h \in \mathcal{E}, r\in \mathcal{R}, t\in \mathcal{E}\}$ is the triple set, where $h$, $r$, and $t$ are the head entity, relation, and tail entity of the triple. 

\begin{figure}[t]
    \centering
    \includegraphics[width=0.45\textwidth]{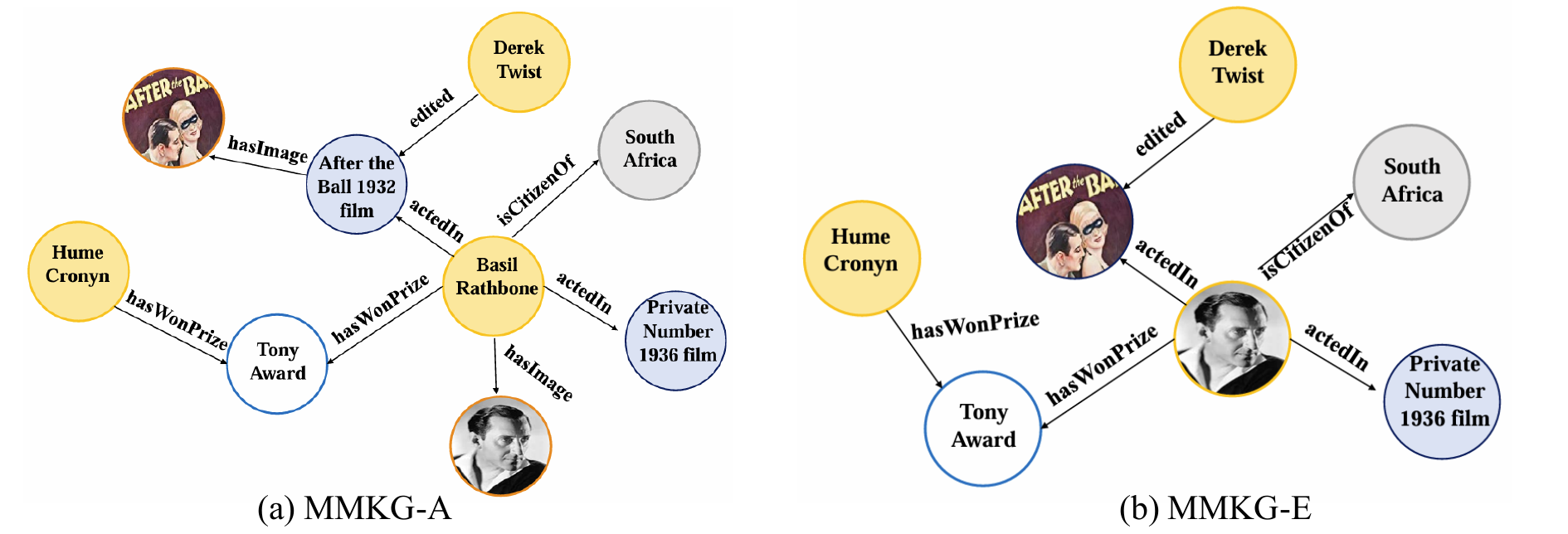}
    \caption{Illustration of multi-modal knowledge graphs}
    \label{fig:MMKG_intro}
\end{figure}

Multi-modal KGs extend KGs by integrating various data types, such as images and text, into a unified knowledge representation. The two primary forms of representation in MMKGs, ``MMKGs-entity type (MMKGs-E)" and ``MMKGs-attribute type (MMKGs-A)" as demonstrated in Fig \ref{fig:MMKG_intro}. MMKGs-E treat multimodal information as independent entities, emphasizing their interactions. In contrast, MMKGs-A view this information as attributes, focusing on the multifaceted characteristics of existing entities. In this paper, we explore generating images for entities in a KG by utilizing its neighbor information, thereby converting the KG into an MMKG-A.

Multi-modal Knowledge Graph Completion (MMKGC) is defined as the process of predicting missing information (such as relations or entities) in a knowledge graph by leveraging data from multiple modalities, including text, images, and structured data. In this work, to specifically explore the impact of images on the completion task, we focus solely on using two modalities: structured data and images.

\section{Related Work}
\subsection{MMKG Construction}
The construction of MMKG has been extensively studied, with a focus primarily on image and structured data sources. Traditional methods often leverage existing multimedia content, such as images, to enrich KGs. For example, 
the OBS MMKG \cite{DBLP:journals/cee/XiongLLL21} is developed by initially identifying entities and relations from heterogeneous, multi-modal data sources. IMGpedia \cite{DBLP:conf/semweb/FerradaBH17} is constructed by extracting relevant visual information from Wikipedia, creating a visual entity for each image, associating visual entities with corresponding Wikipedia articles and establishing links with corresponding entities in DBpedia. 
GAIA \cite{DBLP:conf/acl/LiZLPWCWJCVNF20} introduces a multimedia knowledge extraction system that utilizes visual and text knowledge extraction, and cross-media knowledge fusion and creates a coherent structured knowledge graph. VisualSEM \cite{DBLP:journals/corr/abs-2008-09150} extracts entities, relations, and images from BabelNet v4.0 and applies multiple filtering steps to remove noisy images. Entities are linked to Wikipedia articles, WordNet synsets, and images from ImageNet \cite{DBLP:conf/cvpr/DengDSLL009}. 
This construction approach often involves intricate processes such as the recognition of visual entities/concepts and the fusion across different modalities.

In addition to these repository-based methods, an alternative approach to obtaining images involves leveraging search engines. Projects like ImageGraph \cite{DBLP:conf/akbc/Onoro-RubioNGGL19}, MMKG \cite{DBLP:conf/esws/LiuLGNOR19}, TIVA-KG \cite{wang2023tiva}, and MMpedia \cite{wu2023mmpedia} expand datasets by retrieving images as entities or attributes using search engines. For example, Richpedia \cite{DBLP:journals/bdr/WangWQZ20} filters images from Wikipedia and search engines, while AspectMMKG \cite{DBLP:conf/cikm/ZhangWWLX23} enhances a knowledge graph by collecting feature-specific images and retrieving additional aspect-related images online. Both repository-based and search engine-based methods face challenges in ensuring the selection of high-quality, contextually relevant images for knowledge graph enrichment.

\subsection{Generative AI and Prompt Engineering}
Recent advances in generative AI, such as diffusion models
\cite{DBLP:conf/nips/HoJA20}, DALL-E2 \cite{DBLP:journals/corr/abs-2204-06125}, and Midjourney \footnote{Midjourney. https://www.midjourney.com/, 2023.}, have achieved great success in image generation, opening up new possibilities for enhancing KGs with visual content.  However, current methods that utilize these models often rely on manually crafted prompts to generate images, which can be time-consuming. As the size of the enriched database grows, manually creating prompts becomes increasingly impractical. In response to this challenge, previous studies have explored the use of large language models for automatic prompt generation through techniques such as text mining, text paraphrasing, and data augmentation \cite{DBLP:journals/corr/abs-2304-14670}. \cite{DBLP:conf/wikidata/AhmadCEMRZM23} explored generating images for entities in Wikidata by using the neighbors with the longest token under direct relations with the entity. However, a neighbor's token length does not always indicate its relevance to the entity. 

\section{Method}
\begin{figure*}[ht]
    \centering
    \includegraphics[width=0.90\textwidth]{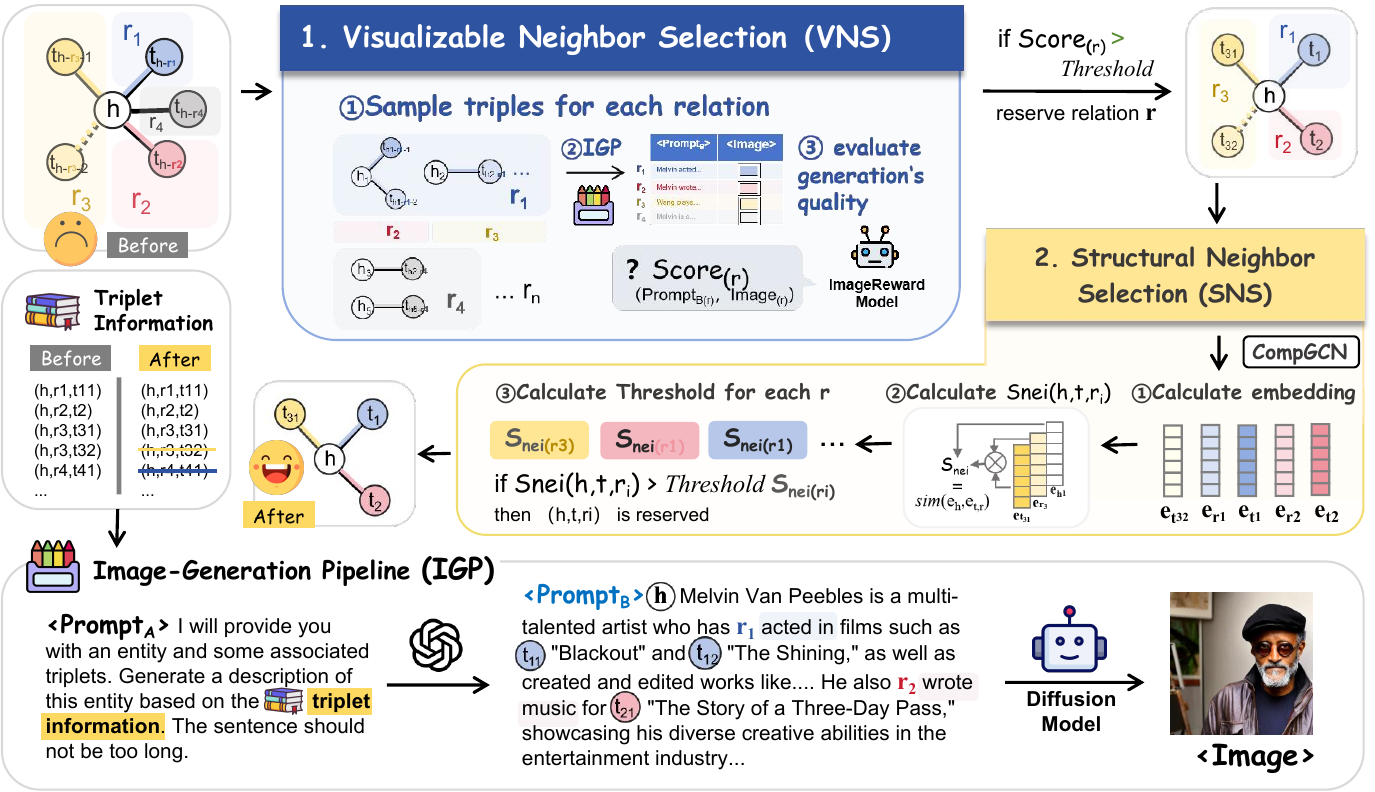} 
    \caption{The framework of translating KGs into MMKGs.}
    \label{fig:model}
\end{figure*}

The framework of our MMKG construction method is illustrated in Fig~\ref{fig:model}. Given a target entity \( e \), the process of generating its description and corresponding image consists of the following three steps:

\begin{enumerate}
    \item \textbf{Visualizable and Structural Neighbor Selection (VSNS):}  
    A set of neighbors \( \mathcal{N}_e \) of \( e \) is selected based on the evaluation of the relations' suitability for visualization. This step ensures that the chosen neighbors are both visually descriptive and structurally relevant to the entity.

    \item \textbf{Semantics-enriched Prompt Generation:}  
    The entity \( e \) and its selected neighbors \( \mathcal{N}_e \) are fed into a Large Language Model (\( \mathcal{LLM} \)) to generate semantics-enriched prompts for \( e \). These prompts capture the contextual and structural information necessary for high-quality image generation.

    \item \textbf{Image Generation:}  
    Using the generated semantics-enriched prompts, a stable diffusion model (\( \mathcal{SD} \)) generates images for the entity \( e \). This step leverages the descriptive power of the prompts to ensure that the generated images are both accurate and visually meaningful.
\end{enumerate}
\subsection{Visualizable and Structural Neighbor Selection}
To generate images for an entity $e$, a straightforward way is to use the name of the entity to generate an image through a text-to-image model such as $\mathcal{SD}$. However, this method may lead to errors or irrelevant images due to the model's limited understanding of the entity's context. Considering the complexity and heterogeneity of KGs and the limited number of input tokens, it is impractical to incorporate all neighbor information into the input for generative models. Moreover, not all relations connected to an entity are suitable for visualization. For instance, functional relations like ``playsFor'' are more visually descriptive than abstract relations like ``influences''. In large-scale KGs, the number of connected entities can range from hundreds to thousands, making it unnecessary to retrieve all neighbors.

To address these challenges, we propose \underline{\textbf{V}}isualizable \underline{\textbf{S}}tructural \underline{\textbf{N}}eighbor \underline{\textbf{S}}election (\textbf{VSNS}), which consists of two modules: \textbf{V}isualizable \textbf{N}eighbor \textbf{S}election (VNS) and \textbf{S}tructural \textbf{N}eighbor \textbf{S}election (SNS). VNS filters relations that are easier to visualize, while SNS selects neighbors that best capture the entity's structural characteristics.

\subsubsection{Visualizable Neighbor Selection} Not all relations connected to an entity contribute to its visualization, and some relations may even hinder visualization \cite{DBLP:conf/wikidata/AhmadCEMRZM23}. 
Given a triple $(h,r,t)$, the VNS module evaluates whether the semantics of \( h \)'s neighbor \( t \) should be considered for \( h \)'s image generation. Specifically, it compares the textual content of each triple associated with \( r \) to the image generated from this text using \textit{ImageReward} \cite{DBLP:conf/nips/XuLWTLDTD23}, a model trained on human preferences to score how well an image represents a text. The visualizability score of a relation \( r \) is calculated as:
\begin{align}
    r_{\text{vis}} = \sum_{i \in \mathcal{T}_r} R_{\text{score}(\text{text}_i,\text{image}_i)} / | \mathcal{T}_r |
\end{align}
where \( \mathcal{T}_r = \{ (e_1, r, e_2) \} \) denotes the selected set of triples associated with \( r \) in the KG, \( \text{text}_i \) represents the textual content generated from the triple \( (h, r, t) \) using a language model \( \mathcal{LM} \), and \( \text{image}_i \) is the corresponding image generated by a stable diffusion model \( \mathcal{SD} \). The function \( R_{\text{score}} \) is defined as:
\begin{align}
    R_{\text{score}(\text{text}_i,\text{image}_i)} = 
    \begin{cases}
        1, & \text{if } \text{ImageReward}(\text{text}_i, \text{image}_i) > 0 \\
        0, & \text{if } \text{ImageReward}(\text{text}_i, \text{image}_i) < 0
    \end{cases}
\end{align}
If \( r_{\text{vis}} \) exceeds a predefined threshold \( \mu \), the neighbors of \( e \) associated with \( r \) are selected for the next image generation step.

\subsubsection{Structural Neighbor Selection}
After applying the VNS module, we obtain a set of triples suitable for visualization. However, in large-scale knowledge graphs, the number of triples connected to an entity \( e \) through a relation \( r \) can still be substantial, ranging from hundreds to thousands. Incorporating all these neighbors as auxiliary information is impractical and unnecessary. Given that entities directly connected are more likely to interact frequently, we aim to select one-hop neighbors based on their structural connectivity.

To capture the structural information in the KG, we pre-train the entire graph using \textit{CompGCN} \cite{DBLP:conf/iclr/VashishthSNT20}, which enriches entity and relation embeddings by integrating both types of information. Let \( \mathbf{e}_v \) and \( \mathbf{e}_r \) denote the embeddings of entity \( v \) and relation \( r \) respectively. For a triple \( (h, r, t) \), the one-hop neighbor representation of \( h \) is computed as:
\begin{align}
    \mathbf{e}_{(r, t)} = \phi_{CompGCN}(\mathbf{e}_{r}, \mathbf{e}_{t})
\end{align}
where \( \phi_{\text{CompGCN}} \) is a composition operator defined by \textit{CompGCN} \cite{DBLP:conf/iclr/VashishthSNT20}. The representation \( \mathbf{e}_{(r, t)} \) captures the structural context of the neighbor \( t \) with respect to the entity \( h \). We then quantify the similarity between \( h \) and its one-hop neighbors using cosine similarity:
\begin{equation}
\mathbf{s}_{nei} = sim(\mathbf{e}_h, \mathbf{e}_{(r,t)}) = \frac{\mathbf{e}_{h}^{\top} \mathbf{e}_{(r,t)}}{\|\mathbf{e}_{h}\| \|\mathbf{e}_{(r,t)}\|}.
\end{equation}

For the neighbors \( t_1, t_2, \dots, t_n \) connected to \( h \) through relation \( r \), we select the neighbor \( t \) that satisfies:
\begin{align}
    \mathbf{s_{nei}}_{t} \ge \sum_{i = 1}^{n} \mathbf{s_{nei}}_{i} / n
\end{align}

\subsection{Semantics-enriched Prompt Generation}
\begin{figure}[t]
    \centering
    \includegraphics[width=0.99\columnwidth]{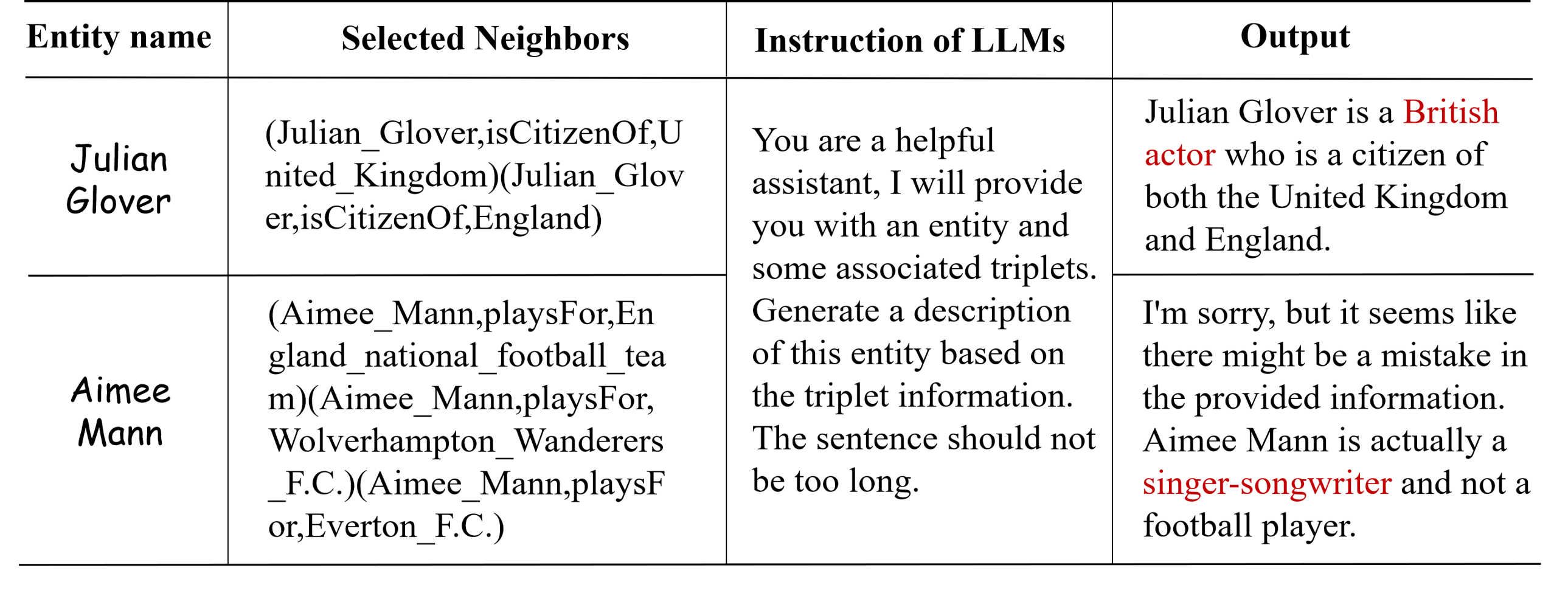}
    \caption{Examples of distilling knowledge from LLM}
    \label{fig:chatgpt_prompt}
\end{figure}

After selecting the relevant neighbors, we leverage a Large Language Model (LLM) to generate descriptive prompts for the entity through specific instructions. This process allows us to distill knowledge from the LLM, enriching the entity's description and verifying the accuracy of neighbor information. For example, as shown in Fig~\ref{fig:chatgpt_prompt}, the LLM added additional information such as "a British actor" for the entity "Julian Glover" and identified inaccuracies in the neighbor information provided for the entity "Aimee Mann" in the knowledge graph.

\subsection{Image Generation}
We employ \textit{Stable Diffusion 2.0-base}\footnote{https://github.com/Stability-AI/stablediffusion} to generate images from the semantics-enriched prompts. This model is chosen for its exceptional ability to capture intricate details and produce realistic images.
\section{Experiment}
\subsection{Experiment Settings}
\paragraph{Datasets}
To evaluate the generalizability of our neighbor selection method across knowledge graphs of varying scales, we conducted experiments on two MMKGs: \textbf{MKG-Y} and \textbf{DB15K}. \textbf{MKG-Y} is a subset of Wikidata \cite{DBLP:journals/cacm/VrandecicK14}, where images are stored as actual image files. \textbf{DB15K} \cite{DBLP:conf/esws/LiuLGNOR19} is derived from Dbpedia \cite{DBLP:journals/semweb/LehmannIJJKMHMK15}, where images are represented as URLs pointing to their respective paths. The detailed statistics of both datasets, including the number of entities, relations, and image references, are summarized in Table \ref{dataset}.
\paragraph{Baselines}
We compare our method against two baselines to demonstrate its effectiveness. The first baseline, denoted as $\text{I}_s$, generates images directly from entity names without leveraging any structural information from the knowledge graph. This approach generates images for all entities indiscriminately. The second baseline, proposed by \cite{DBLP:conf/wikidata/AhmadCEMRZM23}, selects the neighbor with the longest token for each type of connection, focusing exclusively on the head entity. The images generated by this method are denoted as $\text{I}_m$. In contrast, our \textbf{SVNS} method filters neighbors for all entities in the knowledge graph, ensuring a more comprehensive and contextually relevant selection process. The images generated using the filtered neighbors are denoted as $\text{I}_{svns}$.

\begin{table}[ht]
\caption{Statistical information of MKG-Y and DB15K.}
\resizebox{0.47\textwidth}{!}{
\begin{tabular}{ccccccc}
\toprule
\textbf{Dataset} & \textbf{\#Entity} & \textbf{\#Relation} & \textbf{\#Train} & \textbf{\#Valid}                   & \textbf{\#Test} & \textbf{\#Images} \\ \midrule
MKG-Y   & 15000    & 28         & 21310   &2665 & 2663   & 14244    \\
DB15K   & 12842    & 279        & 79222   & 9902                      & 9904   & 12818    \\ \bottomrule
\end{tabular}}
\label{dataset}
\end{table}
\paragraph{Evaluation Metrics}
Given the inherent subjectivity and complexity of determining whether two images depict the same character, we employ a combination of qualitative and quantitative evaluation methods to comprehensively assess the performance of our approach. 
\begin{itemize}
    \item \textbf{Automatic Evaluation.}
    To objectively measure the quality of the generated images, we utilize two widely adopted automated metrics: \textbf{FID} \cite{DBLP:conf/nips/HeuselRUNH17} and \textbf{CLIPscore} \cite{DBLP:conf/emnlp/HesselHFBC21}.
    \begin{itemize}
        \item \textbf{FID} \cite{DBLP:conf/nips/HeuselRUNH17}: 
        This metric evaluates image quality by computing the Fréchet distance between feature distributions of generated and real images, extracted using a pre-trained Inception network. Lower FID values indicate better alignment and realism.
        
        \item \textbf{CLIPscore} \cite{DBLP:conf/emnlp/HesselHFBC21}: 
        This metric measures semantic alignment between images and text by computing the cosine similarity of their embeddings from a contrastive language-image pre-trained model \cite{DBLP:conf/icml/RadfordKHRGASAM21}. It can also assess image-to-image similarity using the CLIP image encoder. We use the CLIP-ViT-B/14 model for robust and accurate evaluation.
    \end{itemize}

    \item \textbf{Human Evaluation.} 
    When generating images for entities in the KG database, our primary objective is to ensure that the generated images accurately reflect the information associated with the respective entities. While automated evaluation metrics, such as FID and CLIPscore, provide quantitative measures of distribution differences and feature similarities between generated and real images, they are often unreliable for determining whether the generated images successfully depict the same characters as the real ones. This is due to potential noise factors, such as variations in image style and color, which can significantly influence the results.

    To address this limitation, we designed a comprehensive manual evaluation questionnaire focusing on three key aspects: \textbf{image quality}, \textbf{entity relevance}, and \textbf{KG relevance}. We recruited three annotators, all with a master's education level, to rank the real images alongside three types of generated images (produced by the two baseline methods and our proposed method). To ensure consistency and minimize inter-annotator variance, we provided the annotators with detailed training prior to the evaluation. 
\end{itemize}
\subsection{Questionnaire Design}
\paragraph{Image Quality (IQ)} 
Each annotator is presented with a randomly selected set of four images, which they evaluate based on image quality. The primary focus on the authenticity of the images, assessed according to the following criteria:
\begin{itemize}
    \item \textbf{Repetitive object generation}: The presence of duplicated objects within the image.
    \item \textbf{Missing limbs}: The absence of limbs or body parts in generated characters.
    \item \textbf{Excessive blurring}: Blurring that significantly impairs the visibility of objects or details.
\end{itemize}
Priority is assigned to images with fewer critical flaws: repetitive generation is considered less severe than missing limbs, which is in turn less severe than excessive blurring. When none of these issues are present, the annotators assign higher scores to images that are clear, aesthetically pleasing, and visually natural.

\paragraph{Correlation between Images and Entity (CIE)} 
Three annotators are tasked with ranking images according to their relevance to a specified entity name. To support this evaluation, we provide additional context by including related neighbors from the KG. This supplementary information enhances the annotators' understanding of the entity, enabling them to make more informed judgments about the relevance of each image to the entity.

\paragraph{Correlation between Images and KG (CIKG)} 
The third questionnaire addresses the challenge of selecting representative images from the potentially large number associated with an entity for integration into the KG. Three annotators are provided with the entity’s name and its KG neighbors, and are tasked with selecting the images that best align with the knowledge graph. This evaluation ensures that the chosen images not only reflect the entity but maintain consistency with the broader context of the KG.

\subsection{Implementation details}
\paragraph{MMKG Generation}
In the VNS module, We employ ImageReward-v1.0\footnote{https://github.com/THUDM/ImageReward} for visual evaluation, where ten triples are sampled per relation and the threshold \(\mu\) is set to 0.5. For the SNS module, we use the "mult" operator from CompGCN \cite{DBLP:conf/iclr/VashishthSNT20}. We utilize ChatGPT as the language model for semantic enrichment. Image features for the generated images are extracted using the CLIP visual encoder \cite{DBLP:conf/icml/RadfordKHRGASAM21}. The MMKG generation pipeline is implemented using PyTorch version 1.8.0 on an Ubuntu 20.04.6 LTS operating system, with computations performed on a single NVIDIA A100 GPU with 40GB of VRAM.
\paragraph{Automatic Evaluation}
Due to the unavailability of many real image URLs in the DB15K dataset (either inaccessible or no longer existing), we focused our FID and CLIPscore calculations exclusively on the MKG-Y dataset. In MKG-Y, each entity with images is associated with three real images. For FID score calculation, we compare the generated image with each of the three real images separately and select the smallest FID value. Similarly, for CLIPscore, we choose the highest score among the comparisons with the real images. 

Given that the MKG-Y training set contains only 7,566 head entities, \cite{DBLP:conf/wikidata/AhmadCEMRZM23} can only generate images with neighbor information for these entities. We conducted evaluations using the two automated metrics on these entities. To ensure meaningful comparisons, we performed pairwise experiments and filtered out entities where different methods selected identical neighbors, as these cases do not provide discriminative insights.

\paragraph{Human Evaluation}
For human evaluation, we randomly sampled 50 entities from the MKG-Y dataset, each associated with ground-truth images. For each entity, we included one real image ($\text{I}_r$), two images generated by the two baseline methods ($\text{I}_s$ and $\text{I}_m$), and one image generated by our VSNS method ($\text{I}_{svns}$). Each sample was evaluated by three annotators, who rated the images on a scale from 1 to 3 based on three criteria: \textbf{Image Quality (IQ)}, \textbf{Correlation between Images and Entity (CIE)}, and \textbf{Correlation between Images and KG (CIKG)}. Higher scores indicate better performance in each criterion. 

Additionally, considering the absence of real images in the DB15K dataset, we randomly sampled 100 entities and performed manual evaluation exclusively on the three types of generated images ($\text{I}_s$, $\text{I}_m$, and $\text{I}_{svns}$). This approach ensures a comprehensive assessment of the generated images across both datasets.
\begin{table}[ht]
\centering
\caption{The results of FID and CLIPscore on MKG-Y}
\renewcommand{\arraystretch}{1.07} 
\resizebox{0.45\textwidth}{!}{
\begin{tabular}{cllc}
\toprule
\textbf{Methods} & \textbf{FID}\,($\downarrow$)  & \textbf{CLIPscore}\,($\uparrow$) & \textbf{\#C\_Number}           \\ \midrule
$\text{I}_s$      & 280.6 & \quad0.619     & \multirow{2}{*}{6409} \\
\quad\ \,$\text{I}_{svns}$   & 272.7\textcolor{deepgreen}{\,\tiny$\downarrow7.9$} &\quad 0.647 \textcolor{deepgreen}{\,\tiny$\uparrow0.028$}    &                       \\ \midrule
$\text{I}_m$      & 275.7 &\quad 0.659     & \multirow{2}{*}{3643} \\
\quad$\text{I}_{svns}$  & 267.7\textcolor{deepgreen}{\,\tiny$\downarrow8.0$} &\quad 0.666 \textcolor{deepgreen}{\,\tiny$\uparrow0.007$}    &                       \\ \bottomrule
\end{tabular}}
\label{automatic_results}
\end{table}


\begin{figure}[htbp]
  \centering
  \includegraphics[width=0.99\columnwidth]{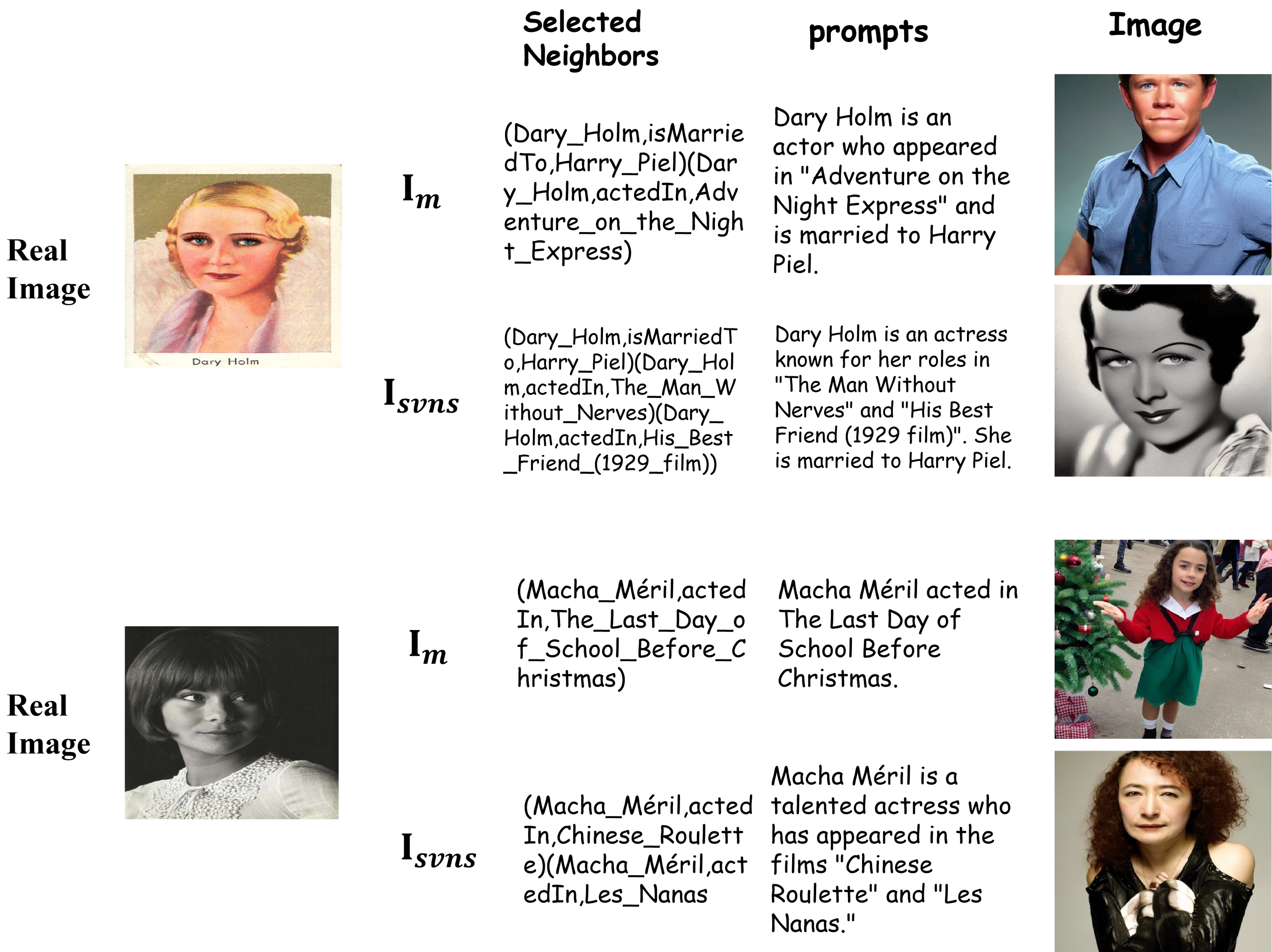}
  \caption{Example of $\text{I}_m$ with lower CIE.}
  \label{fig:maxtoken_case_bad}
\end{figure}
\subsection{Main Results}
\paragraph{Automatic Evaluation}
To ensure a fair comparison between the VSNS method and the two baselines, we filtered out entities where identical neighbors were selected by both methods. The final number of entities compared, along with the calculated results for the two metrics, are presented in Table \ref{automatic_results}. The results show that the SVNS method achieves superior performance compared to the baseline methods on all evaluation metrics, producing images of higher quality that exhibit greater similarity to the features of real images.

For the set of 6,409 entities, $\text{I}_{svns}$ achieved an average FID of 272.7, which is lower than the 280.6 achieved by $\text{I}_s$. Similarly, for the set of 3,643 entities, $\text{I}_{svns}$ attained an average FID of 267.7, outperforming the 275.7 of $\text{I}_m$. These results indicate that SVNS generates images with significantly higher visual fidelity compared to the baseline methods.

Furthermore, SVNS excels in terms of average CLIPscore performance. The average CLIPscore for $\text{I}_{svns}$ was 0.647 and 0.666 for the two sets of entities, representing improvements of $2.8\%$ and $0.7\%$ over $\text{I}_s$ and $\text{I}_m$, respectively. The higher CLIPscore values underscore SVNS's ability to produce images that are more semantically aligned with the ground-truth images, further validating its effectiveness.

\paragraph{Human Evaluation}
The results of the human evaluation are presented in Table \ref{Human_evaluation}. The table indicates that images generated using neighbors filtered by the SVNS method exhibit the highest relevance to the KG among all generated images. In terms of both image quality and entity relevance, $I_{svns}$ outperforms $I_m$. While the generated images still show some gaps in image quality and entity relevance compared to real images, they demonstrate strong performance in knowledge graph adaptability, highlighting their potential for specific tasks.

\begin{figure*}[htbp]
  \centering
  \includegraphics[scale=0.40]{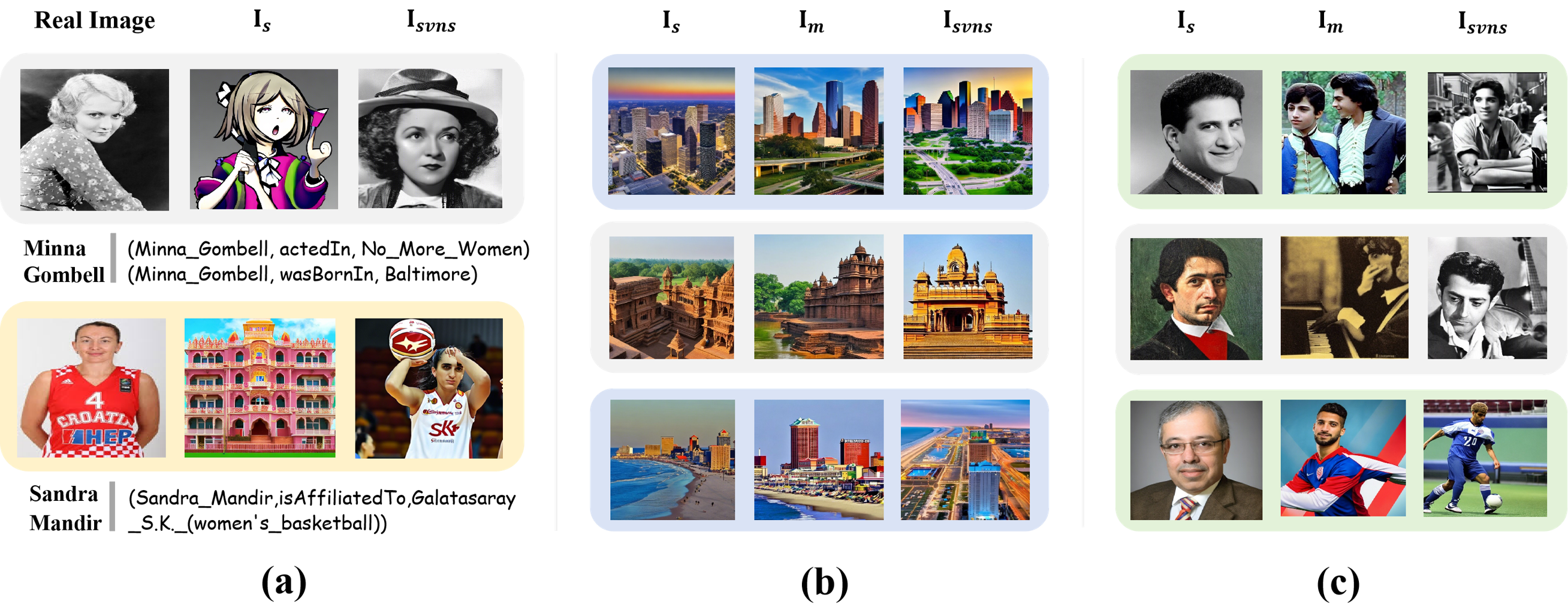}
  \caption{(a): example of $\text{I}_s$ with lower CIE. (b): examples of generated images of landscapes or
places. (c): example of $\text{I}_{svns}$ or $\text{I}_m$ with lower IQ.}
  \label{fig:case}
\end{figure*}

We further analyzed the reasons for the lower scores in image quality for images generated using neighbor information in the MKG-Y dataset. The primary issues identified were incomplete body parts and repetitive generation when creating characters, as illustrated in Fig \ref{fig:case}(c). However, this does not imply that the images generated by $\text{I}_s$ are representative of the entity.

For entity relevance scores, we examined the lower-scoring images generated by the two baseline methods ($\text{I}_s$ and $\text{I}_m$). For $\text{I}_s$, the main issue was the lack of specific entity information, resulting in generated images that do not align with the entities in the knowledge graph. Specific examples are shown in Fig \ref{fig:case}(a). For $\text{I}_m$, since \cite{DBLP:conf/wikidata/AhmadCEMRZM23} selects neighbors based on token length, longer token lengths do not necessarily indicate stronger relevance to the entity. As shown in Fig \ref{fig:maxtoken_case_bad}, for the actor "Dary Holm", the movie "Adventure on the Night Express", selected by  \cite{DBLP:conf/wikidata/AhmadCEMRZM23}, is actually a representative work of the actor ''Harry Piel", not "Dary Holm". In contrast, the SVNS method successfully identified "The Man Without Nerves" as a representative work for "Dary Holm", resulting in images with higher relevance to the entity.

Additionally, we observed that when generating images for landscapes or places, there were no significant differences among $\text{I}_s$, $\text{I}_m$, and $\text{I}_{svns}$, as illustrated in Fig \ref{fig:case}(b).
\begin{table}[ht]
\renewcommand{\arraystretch}{1.07} 
\centering
\caption{The human evaluation results on MKG-Y and DB15K.}
\resizebox{0.45\textwidth}{!}{
\begin{tabular}{lcccc|ccc}
\toprule
\multirow{2}{*}{\textbf{Criteria}\quad} & \multicolumn{4}{c}{\textbf{MKG-Y}}  & \multicolumn{3}{c}{\textbf{DB15K}} \\ \cmidrule{2-8} 
                  & $\text{I}_r$ & $\text{I}_s$ & $\text{I}_m$ & $\text{I}_{svns}$ & $\text{I}_s$   & $\text{I}_m$  & $\text{I}_{svns}$  \\ \midrule
IQ                & 3.33 & 2.72 & 2.36 & 2.68    & 2.06   & 1.88  & 2.26      \\
CIE               & 3.90 & 2.00 & 2.06 & 2.27    & 2.31   & 2.07  & 2.27      \\ 
CIKG              & 2.32 & 2.06 & 2.24 & 2.48    & 1.99   & 2.16  & 2.44       \\ \bottomrule
\end{tabular}}
\label{Human_evaluation}
\end{table} 
\subsection{Ablation study}
\paragraph{The Impact of the VNS Module}
To validate the effectiveness of the VNS module, we analyzed the change in image quality for entities that originally contained only a single relation \( r \). Specifically, we focused on entities with only one relation that were affected by the VNS module. The results, as shown in Table \ref{vns_analysis}, demonstrate that the VNS module significantly improves the quality of generated images.

When the VNS module is applied, the FID decreases from 359.0 to 349.3, while the CLIPscore increases from 0.554 to 0.559. These improvements indicate enhanced image quality and feature similarity compared to not using the VNS module. The impact of the VNS module becomes even more pronounced when compared to images generated by the baseline method ($\text{I}_m$), with a substantial reduction in FID (from 303.7 to 269.8) and an increase in CLIPscore (from 0.625 to 0.642). These results underscore the effectiveness of the VNS module in improving both image quality and semantic alignment. Overall, the VNS module consistently outperforms the baseline methods across all evaluation metrics.
\begin{table}[]
\centering
\caption{The impact analysis of the VNS module.}
\renewcommand{\arraystretch}{1.07} 
\resizebox{0.45\textwidth}{!}{
\begin{tabular}{cccc}
\toprule
\multicolumn{1}{c}{\textbf{Methods}} & \multicolumn{1}{c}{\textbf{FID}\,($\downarrow$)} & \textbf{CLIPscore}\,($\uparrow$) & \textbf{\#C\_Number}           \\ \midrule
$\text{I}_{wo\_vns}$                   & 359.0                    & 0.554     & \multirow{2}{*}{3613} \\
$\text{I}_{w\_vns}$                    & 349.3                    & 0.559     &                       \\ \midrule
$\text{I}_{wo\_vns}$                   & 303.7                    & 0.625     & \multirow{3}{*}{272}  \\
$\text{I}_m$                         & 310.7                    & 0.629     &                       \\
$\text{I}_{w\_vns}$                    & \textbf{269.8}                    & \textbf{0.642}     &                       \\ \bottomrule
\end{tabular}}
\label{vns_analysis}
\end{table}
\begin{table}[ht]
\centering
\caption{The impact analysis of the SNS module.}
\renewcommand{\arraystretch}{1.07} 
\resizebox{0.45\textwidth}{!}{
\begin{tabular}{cccc}
\toprule
\multicolumn{1}{c}{\textbf{Methods}} & \multicolumn{1}{c}{\textbf{FID}\,($\downarrow$)} & \textbf{CLIPscore}\,($\uparrow$) & \textbf{\#C\_Number}           \\ \midrule
$\text{I}_s$                         & 280.7                    & 0.615     & \multirow{3}{*}{4509} \\ 
$\text{I}_m $                        & 269.7                    & 0.644     &                       \\ 
$\text{I}_{sns}$                       & \textbf{266.1}                    & \textbf{0.650}     &                       \\ \bottomrule
\end{tabular}}
\vspace{-2mm}
\label{sns_analyse}
\end{table}
\paragraph{The Impact of the SNS Module}
The primary objective of the SNS module is to filter the neighbors of entities effectively. To evaluate its performance in complex scenarios, we focused on entities with more than one neighbor. The results, as shown in Table \ref{sns_analyse}, demonstrate that the SNS module significantly improves both the quality and feature similarity of generated images.

Compared to the two baseline methods, the use of the SNS module ($\text{I}_{sns}$) achieves the lowest FID score of 266.1 and the highest CLIPscore of 0.650. These results indicate that the generated images are closer to the distribution of real images and exhibit stronger feature similarity. Overall, the SNS module proves to be highly effective in enhancing the quality of image generation, particularly in complex scenarios with multiple neighbors.

\begin{table}[ht]
\renewcommand{\arraystretch}{1.07} 
\caption{The results of MMKGC on MKG-Y and DB15K. The \textbf{bold} represents the best results.}
\centering
\resizebox{0.45\textwidth}{!}{
\begin{tabular}{c|c|cc|cc}
\toprule
\multicolumn{1}{c}{\multirow{2}{*}{\textbf{Modality}}} & \multicolumn{1}{c}{\multirow{2}{*}{\textbf{Methods}}} & \multicolumn{2}{c}{\textbf{MKG-Y}} & \multicolumn{2}{c}{\textbf{DB15K}} \\ \cmidrule{3-6}
 \multicolumn{1}{c}{}                          & \multicolumn{1}{c}{}                         & MRR          & \multicolumn{1}{c}{Hit@1}        & MRR         & Hit@1       \\ \midrule
S                         & TransE                   & 0.307        & 0.235        & 0.249       & 0.128       \\
S                         & RotatE                   & 0.350        & 0.291        & 0.293       & 0.179       \\
S                         & PairRE                   & 0.320        & 0.256        & 0.311       & 0.216       \\ \midrule
S+I                       & NATIVE                   & 0.383        & \textbf{0.346}        & 0.355       & 0.269       \\
\quad \;\,\  S+$\text{I}_{svns}$                 & NATIVE                   & \textbf{0.387}     & 0.345        & \textbf{0.368}       & \textbf{0.276}       \\ \bottomrule
\end{tabular}}
\label{MMKGC}
\end{table}
\paragraph{The impact of generating images}
To investigate the potential benefits of generated images for downstream tasks in knowledge graphs, we focused on the widely used knowledge graph reasoning task—Knowledge Graph Completion (KGC). To evaluate the impact of generated images on KGC, we employed the state-of-the-art NATIVE method \cite{DBLP:conf/sigir/ZhangCGXHLZC24}, which integrates both generated images and the original structural information of the graph in a multimodal knowledge graph completion (MMKGC) experiment. The evaluation metrics used are MRR (Mean Reciprocal Rank) and Hits@N, which are standard in link prediction tasks. Higher values for these metrics indicate better performance, with MRR reflecting the average quality of rankings and Hits@N measuring the proportion of correct predictions within the top-N ranked results.

We conducted experiments on the MKG-Y and DB15K datasets, and the results are summarized in Table \ref{MMKGC}. The findings reveal that incorporating image information significantly boosts the performance of knowledge graph reasoning tasks. Compared to using only structural information (S), the addition of real images (S+I) leads to substantial improvements, particularly in MRR and Hits@1 metrics. Notably, when comparing generated images (S+$\text{I}_{svns}$) to real images (S+I), the generated images in certain cases achieve comparable or even superior performance, especially in the DB15K dataset. This indicates that, although synthetic, the generated images provide meaningful feature information for multimodal knowledge graph reasoning tasks, highlighting the potential of the SVNS method to produce image features that are competitive with those derived from real images.

\section{Conclusions and Future Work}
In this paper, we introduce a framework for transforming KGs into MMKGs by generating entity-specific images. To enhance the relevance of these images to both the entities and the information in the knowledge graph, we designed a neighbor selection method named VSNS. VSNS contains two core modules: VNS, which filters out relations that are challenging to visualize, and SNS,  which selects neighbors that better represent the structural attributes of the entities. Quantitative and qualitative experiments have demonstrated the superiority of our method. Current limitations include handling abstract entities (e.g., emotions, events) and unvalidated downstream performance (e.g., QA, recommendations). Future work will address these gaps.
\section*{Acknowledgment}
This work is founded by National Natural Science Foundation of China (NSFCU23B2055 / NSFCU19B2027 / NSFC62306276), Zhejiang Provincial Natural Science Foundation of China (No. LQ23F020017), Yongjiang Talent Introduction Programme (2022A-238-G), and Fundamental Research Funds for the Central Universities (226-2023-00138). This work was supported by AntGroup.
\bibliographystyle{IEEEtran}
\bibliography{aaai25}

\begin{thebibliography}{10}
\providecommand{\url}[1]{#1}
\csname url@samestyle\endcsname
\providecommand{\newblock}{\relax}
\providecommand{\bibinfo}[2]{#2}
\providecommand{\BIBentrySTDinterwordspacing}{\spaceskip=0pt\relax}
\providecommand{\BIBentryALTinterwordstretchfactor}{4}
\providecommand{\BIBentryALTinterwordspacing}{\spaceskip=\fontdimen2\font plus
\BIBentryALTinterwordstretchfactor\fontdimen3\font minus \fontdimen4\font\relax}
\providecommand{\BIBforeignlanguage}[2]{{%
\expandafter\ifx\csname l@#1\endcsname\relax
\typeout{** WARNING: IEEEtran.bst: No hyphenation pattern has been}%
\typeout{** loaded for the language `#1'. Using the pattern for}%
\typeout{** the default language instead.}%
\else
\language=\csname l@#1\endcsname
\fi
#2}}
\providecommand{\BIBdecl}{\relax}
\BIBdecl

\bibitem{chen2024knowledge}
Z.~Chen, Y.~Zhang, Y.~Fang, Y.~Geng, L.~Guo, X.~Chen, Q.~Li, W.~Zhang, J.~Chen, Y.~Zhu \emph{et~al.}, ``Knowledge graphs meet multi-modal learning: A comprehensive survey,'' \emph{arXiv preprint arXiv:2402.05391}, 2024.

\bibitem{DBLP:journals/tgdk/PanRKSCDJO0LBMB23}
J.~Z. Pan, S.~Razniewski, J.~Kalo, S.~Singhania, J.~Chen, S.~Dietze, H.~Jabeen, J.~Omeliyanenko, W.~Zhang, M.~Lissandrini, R.~Biswas, G.~de~Melo, A.~Bonifati, E.~Vakaj, M.~Dragoni, and D.~Graux, ``Large language models and knowledge graphs: Opportunities and challenges,'' \emph{{TGDK}}, vol.~1, no.~1, pp. 2:1--2:38, 2023.

\bibitem{DBLP:conf/nips/HedlinSMIKTY23}
E.~Hedlin, G.~Sharma, S.~Mahajan, H.~Isack, A.~Kar, A.~Tagliasacchi, and K.~M. Yi, ``Unsupervised semantic correspondence using stable diffusion,'' in \emph{NeurIPS}, 2023.

\bibitem{DBLP:conf/nips/HoJA20}
J.~Ho, A.~Jain, and P.~Abbeel, ``Denoising diffusion probabilistic models,'' in \emph{NeurIPS}, 2020.

\bibitem{DBLP:conf/wikidata/AhmadCEMRZM23}
R.~A. Ahmad, M.~Critelli, S.~Efeoglu, E.~Mancini, C.~Ringwald, X.~Zhang, and A.~Mero{\~{n}}o{-}Pe{\~{n}}uela, ``Draw me like my triples: Leveraging generative {AI} for wikidata image completion,'' in \emph{Wikidata@ISWC}, ser. {CEUR} Workshop Proceedings, vol. 3640.\hskip 1em plus 0.5em minus 0.4em\relax CEUR-WS.org, 2023.

\bibitem{DBLP:journals/cee/XiongLLL21}
J.~Xiong, G.~Liu, Y.~Liu, and M.~Liu, ``Oracle bone inscriptions information processing based on multi-modal knowledge graph,'' \emph{Comput. Electr. Eng.}, vol.~92, p. 107173, 2021.

\bibitem{DBLP:conf/semweb/FerradaBH17}
S.~Ferrada, B.~Bustos, and A.~Hogan, ``Imgpedia: {A} linked dataset with content-based analysis of wikimedia images,'' in \emph{{ISWC} {(2)}}, ser. Lecture Notes in Computer Science, vol. 10588.\hskip 1em plus 0.5em minus 0.4em\relax Springer, 2017, pp. 84--93.

\bibitem{DBLP:conf/acl/LiZLPWCWJCVNF20}
M.~Li, A.~Zareian, Y.~Lin, X.~Pan, S.~Whitehead, B.~Chen, B.~Wu, H.~Ji, S.~Chang, C.~R. Voss, D.~Napierski, and M.~Freedman, ``{GAIA:} {A} fine-grained multimedia knowledge extraction system,'' in \emph{{ACL} (demo)}.\hskip 1em plus 0.5em minus 0.4em\relax Association for Computational Linguistics, 2020, pp. 77--86.

\bibitem{DBLP:journals/corr/abs-2008-09150}
H.~Alberts, T.~Huang, Y.~Deshpande, Y.~Liu, K.~Cho, C.~Vania, and I.~Calixto, ``Visualsem: a high-quality knowledge graph for vision and language,'' \emph{CoRR}, vol. abs/2008.09150, 2020.

\bibitem{DBLP:conf/cvpr/DengDSLL009}
J.~Deng, W.~Dong, R.~Socher, L.~Li, K.~Li, and L.~Fei{-}Fei, ``Imagenet: {A} large-scale hierarchical image database,'' in \emph{{CVPR}}.\hskip 1em plus 0.5em minus 0.4em\relax {IEEE} Computer Society, 2009, pp. 248--255.

\bibitem{DBLP:conf/akbc/Onoro-RubioNGGL19}
D.~O{\~{n}}oro{-}Rubio, M.~Niepert, A.~Garc{\'{\i}}a{-}Dur{\'{a}}n, R.~Gonzalez{-}Sanchez, and R.~J. L{\'{o}}pez{-}Sastre, ``Answering visual-relational queries in web-extracted knowledge graphs,'' in \emph{{AKBC}}, 2019.

\bibitem{DBLP:conf/esws/LiuLGNOR19}
Y.~Liu, H.~Li, A.~Garc{\'{\i}}a{-}Dur{\'{a}}n, M.~Niepert, D.~O{\~{n}}oro{-}Rubio, and D.~S. Rosenblum, ``{MMKG:} multi-modal knowledge graphs,'' in \emph{{ESWC}}, ser. Lecture Notes in Computer Science, vol. 11503.\hskip 1em plus 0.5em minus 0.4em\relax Springer, 2019, pp. 459--474.

\bibitem{wang2023tiva}
X.~Wang, B.~Meng, H.~Chen, Y.~Meng, K.~Lv, and W.~Zhu, ``{TIVA-KG:} {A} multimodal knowledge graph with text, image, video and audio,'' in \emph{{ACM} Multimedia}.\hskip 1em plus 0.5em minus 0.4em\relax {ACM}, 2023, pp. 2391--2399.

\bibitem{wu2023mmpedia}
Y.~Wu, X.~Wu, J.~Li, Y.~Zhang, H.~Wang, W.~Du, Z.~He, J.~Liu, and T.~Ruan, ``Mmpedia: A large-scale multi-modal knowledge graph,'' in \emph{ISWC}.\hskip 1em plus 0.5em minus 0.4em\relax Springer, 2023, pp. 18--37.

\bibitem{DBLP:journals/bdr/WangWQZ20}
M.~Wang, H.~Wang, G.~Qi, and Q.~Zheng, ``Richpedia: {A} large-scale, comprehensive multi-modal knowledge graph,'' \emph{Big Data Res.}, vol.~22, p. 100159, 2020.

\bibitem{DBLP:conf/cikm/ZhangWWLX23}
J.~Zhang, J.~Wang, X.~Wang, Z.~Li, and Y.~Xiao, ``Aspectmmkg: {A} multi-modal knowledge graph with aspect-aware entities,'' in \emph{{CIKM}}.\hskip 1em plus 0.5em minus 0.4em\relax {ACM}, 2023, pp. 3361--3370.

\bibitem{DBLP:journals/corr/abs-2204-06125}
A.~Ramesh, P.~Dhariwal, A.~Nichol, C.~Chu, and M.~Chen, ``Hierarchical text-conditional image generation with {CLIP} latents,'' \emph{CoRR}, vol. abs/2204.06125, 2022.

\bibitem{DBLP:journals/corr/abs-2304-14670}
J.~Wang, E.~Shi, S.~Yu, Z.~Wu, C.~Ma, H.~Dai, Q.~Yang, Y.~Kang, J.~Wu, H.~Hu, C.~Yue, H.~Zhang, Y.~Liu, X.~Li, B.~Ge, D.~Zhu, Y.~Yuan, D.~Shen, T.~Liu, and S.~Zhang, ``Prompt engineering for healthcare: Methodologies and applications,'' \emph{CoRR}, vol. abs/2304.14670, 2023.

\bibitem{DBLP:conf/nips/XuLWTLDTD23}
J.~Xu, X.~Liu, Y.~Wu, Y.~Tong, Q.~Li, M.~Ding, J.~Tang, and Y.~Dong, ``Imagereward: Learning and evaluating human preferences for text-to-image generation,'' in \emph{NeurIPS}, 2023.

\bibitem{DBLP:conf/iclr/VashishthSNT20}
S.~Vashishth, S.~Sanyal, V.~Nitin, and P.~P. Talukdar, ``Composition-based multi-relational graph convolutional networks,'' in \emph{{ICLR}}.\hskip 1em plus 0.5em minus 0.4em\relax OpenReview.net, 2020.

\bibitem{DBLP:journals/cacm/VrandecicK14}
D.~Vrandecic and M.~Kr{\"{o}}tzsch, ``Wikidata: a free collaborative knowledgebase,'' \emph{Commun. {ACM}}, vol.~57, no.~10, pp. 78--85, 2014.

\bibitem{DBLP:journals/semweb/LehmannIJJKMHMK15}
J.~Lehmann, R.~Isele, M.~Jakob, A.~Jentzsch, D.~Kontokostas, P.~N. Mendes, S.~Hellmann, M.~Morsey, P.~van Kleef, S.~Auer, and C.~Bizer, ``Dbpedia - {A} large-scale, multilingual knowledge base extracted from wikipedia,'' \emph{Semantic Web}, vol.~6, no.~2, pp. 167--195, 2015.

\bibitem{DBLP:conf/nips/HeuselRUNH17}
M.~Heusel, H.~Ramsauer, T.~Unterthiner, B.~Nessler, and S.~Hochreiter, ``Gans trained by a two time-scale update rule converge to a local nash equilibrium,'' in \emph{{NIPS}}, 2017, pp. 6626--6637.

\bibitem{DBLP:conf/emnlp/HesselHFBC21}
J.~Hessel, A.~Holtzman, M.~Forbes, R.~L. Bras, and Y.~Choi, ``Clipscore: {A} reference-free evaluation metric for image captioning,'' in \emph{{EMNLP} {(1)}}.\hskip 1em plus 0.5em minus 0.4em\relax Association for Computational Linguistics, 2021, pp. 7514--7528.

\bibitem{DBLP:conf/icml/RadfordKHRGASAM21}
A.~Radford, J.~W. Kim, C.~Hallacy, A.~Ramesh, G.~Goh, S.~Agarwal, G.~Sastry, A.~Askell, P.~Mishkin, J.~Clark, G.~Krueger, and I.~Sutskever, ``Learning transferable visual models from natural language supervision,'' in \emph{{ICML}}, ser. Proceedings of Machine Learning Research, vol. 139.\hskip 1em plus 0.5em minus 0.4em\relax {PMLR}, 2021, pp. 8748--8763.

\bibitem{DBLP:conf/sigir/ZhangCGXHLZC24}
Y.~Zhang, Z.~Chen, L.~Guo, Y.~Xu, B.~Hu, Z.~Liu, W.~Zhang, and H.~Chen, ``Native: Multi-modal knowledge graph completion in the wild,'' in \emph{{SIGIR}}.\hskip 1em plus 0.5em minus 0.4em\relax {ACM}, 2024, pp. 91--101.

\end{thebibliography}

\end{document}